\documentclass[10pt,twocolumn,letterpaper]{article}
\usepackage{titlesec}

\usepackage{cvpr}
\usepackage{times}
\usepackage{epsfig}
\usepackage{graphicx}
\usepackage{amsmath}
\usepackage{amssymb}
\usepackage{multirow}
\usepackage{booktabs}

\usepackage[pagebackref=true,breaklinks=true,letterpaper=true,colorlinks,bookmarks=false]{hyperref}

\cvprfinalcopy 

\def\cvprPaperID{2735} 

\ifcvprfinal\fi
\usepackage{fancyvrb}

\titlespacing\section{0pt}{12pt plus 1pt minus 2pt}{0pt plus 1pt minus 2pt}
\titlespacing\subsection{0pt}{12pt plus 1pt minus 2pt}{0pt plus 1pt minus 2pt}
\titlespacing\subsubsection{0pt}{12pt plus 1pt minus 2pt}{0pt plus 1pt minus 2pt}

\begin{document}

\title{DecideNet: Counting Varying Density Crowds \\
	Through Attention Guided Detection and Density Estimation}

\author{Jiang Liu$^1$, Chenqiang Gao$^2$, Deyu Meng$^3$, Alexander G. Hauptmann$^1$\\
$^1$Carnegie Mellon University \\
$^2$Chongqing University of Posts and Telecommunications 
$^3$Xi'an Jiaotong University \\
{\tt\small $^1$\{jiangl1,alex\}@cs.cmu.edu, \tt\small $^2$gaocq@cqupt.edu.cn, \tt\small $^3$dymeng@mail.xjtu.edu.cn}
}

\maketitle

\begin{abstract}
In real-world crowd counting applications, the crowd densities vary greatly in spatial and temporal domains.
A detection based counting method will estimate crowds accurately in low density scenes, while its reliability in congested areas is downgraded.
A regression based approach, on the other hand, captures the general density information in crowded regions.
Without knowing the location of each person, it tends to overestimate the count in low density areas.
Thus, exclusively using either one of them is not sufficient to handle all kinds of scenes with varying densities.
To address this issue, a novel end-to-end crowd counting framework, named DecideNet (DEteCtIon and Density Estimation Network) is proposed. 
It can adaptively decide the appropriate counting mode for different locations on the image based on its real density conditions.
DecideNet starts with estimating the crowd density by generating detection and regression based density maps separately.
To capture inevitable variation in densities, it incorporates an attention module, meant to adaptively assess the reliability of the two types of estimations.
The final crowd counts are obtained with the guidance of the attention module to adopt suitable estimations from the two kinds of density maps. 
Experimental results show that our method achieves state-of-the-art performance on three challenging crowd counting datasets.

\end{abstract}

\section{Introduction}
The crowd counting task in the computer vision community aims at obtaining number of individuals appearing in specific scenes.
It is the essential building block for high-level crowd analysis, including crowd monitoring \cite{chan2008privacy}, scene understanding \cite{zhang2015cross} and public safety management \cite{chen2016videos}.
\begin{figure}[h]
	\centering
	\includegraphics[width=7.5cm]{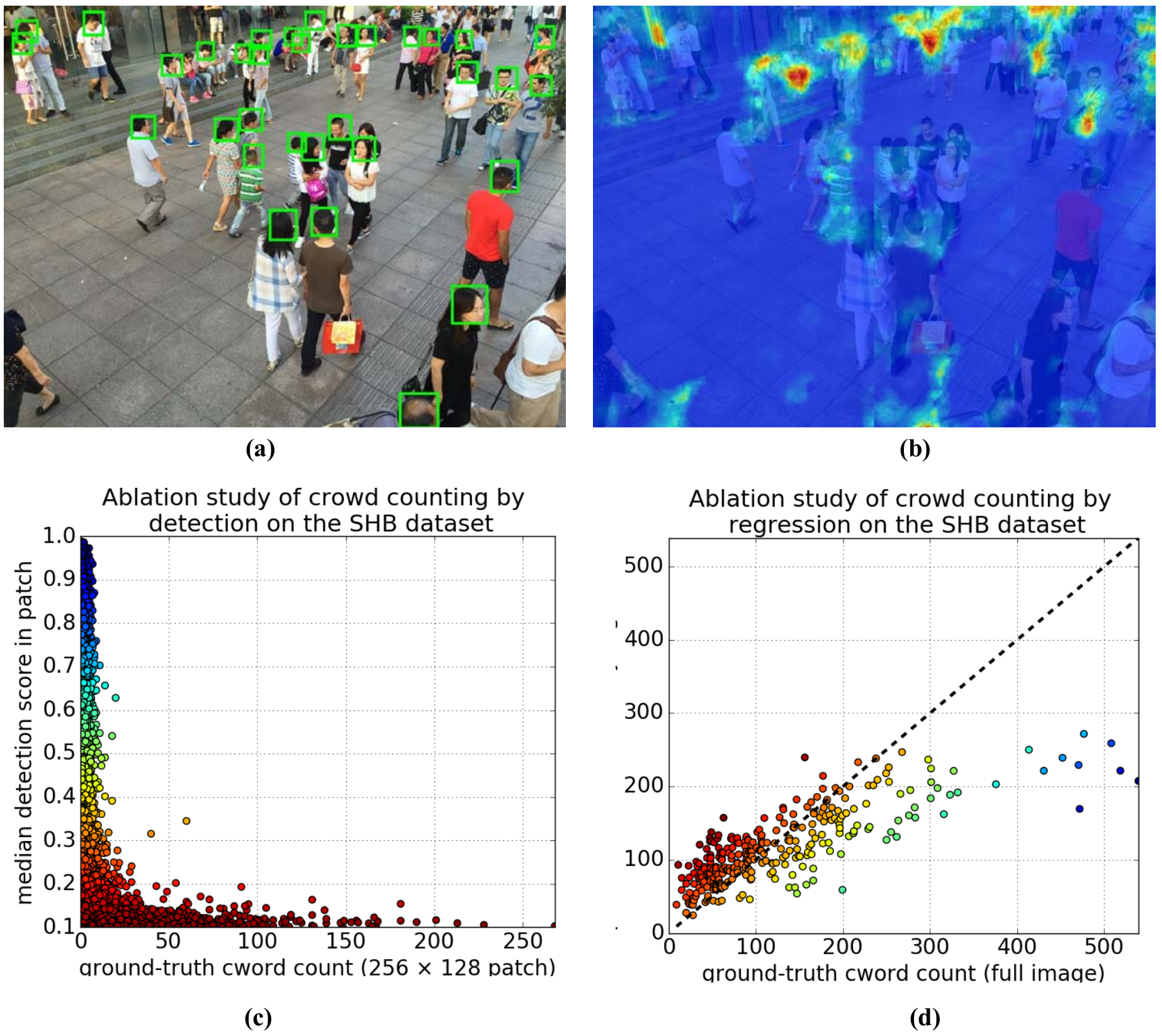}
	\caption{Ablation studies of detection and regression based crowd counting on the ShanghaiTech PartB (SHB) dataset \cite{zhang2016single}. 
		Detection reliability decreases along with the increased crowd density, resulting in \emph{underestimated} counts in those areas.
		Counts from density estimation tend to be \emph{overestimated} in scenes with low densities.
		(a) Visualization of the detection results on a image from a Faster R-CNN \cite{ren2015faster} detector. 
		(b) The density map on the same image from a CNN regression network \cite{sam2017switching}. 
		(c) The median object detection scores from the detector used in (a) versus the ground-truth counts. 
		(d) The predictions from the network used in (b) versus the true crowd counts.}
	\label{fig:fig_det_reg} 
\end{figure}

Various methods have been proposed to tackle this problem. 
They could generally be classified into detection and regression based approaches.
The detection based methods \cite{dalal2005histograms,leibe2005pedestrian,viola2003detecting,felzenszwalb2010object,zhao2008segmentation,ge2009marked} employ object detectors to localize the position for each person. 
The number of detections is then treated as the crowd count. 
Early works \cite{dalal2005histograms,viola2003detecting,felzenszwalb2010object} employ low-level features as region descriptors, followed by a classifier for classification.
Benefiting from the recent progress in object detection using deep neural networks \cite{girshick2015fast,ren2015faster,redmon2016you,gao2016people}, in ideal images with relatively large individual sizes and sparse crowd densities, detection based counting could surpass human performance. 
Different from crowd counting by detection, regression based methods \cite{lempitsky2010learning,pham2015count,xu2016crowd,sam2017switching,zhang2016single,boominathan2016crowdnet} obtain the crowd count without explicitly detecting and localizing each individual.
Preliminary works directly learn the mapping between features of image patches to crowd counts \cite{lempitsky2010learning,pham2015count,xu2016crowd}.
Recent regression based works improve the performance with Convolutional Neural Network (CNN) \cite{boominathan2016crowdnet,sam2017switching,zhang2016single,marsden2016fully,onoro2016towards} to output density maps of image patches. 
Integrating over the map will give the count for the patch.
Regression based methods usually work well in crowded patches since they can capture the general density information by benefiting from the rich context in local patches.

In real-world counting applications, the crowd density varies enormously in spatial and temporal domains.
In the spatial aspect, even in a same image, the density in some regions may be much higher than those of others.
In some background regions, there may even be no person present. 
Meanwhile, it is also natural for the crowd volume to change along with time: a business street  may have very high crowd volumes during the workdays, while the weekends counterparts are much lower.
Intuitively, here comes a question: \emph{can crowd counting exclusively based on either regression or detection be enough to simultaneously handle high and low density scenes?}

To answer this question, we study the performance of two types of approaches on the ShanghaiTech PartB (SHB) dataset \cite{zhang2016single} collected from real street scenes with great variation in crowd densities.
The result is illustrated in Figure \ref{fig:fig_det_reg}.
Figure \ref{fig:fig_det_reg}(a) gives the detections from a fine-tuned Faster R-CNN head detector on a specific image: with the distance to the camera increasing, the crowd density and the number of missed detections rises.
Figure \ref{fig:fig_det_reg}(c) shows the relationship between median detection scores and  ground-truth counts for 10,000 image patches with sizes of $256 \times 128$.
It is clear that the score drops rapidly with the rise of the ground-truth count.
We may therefore find that the reliability of detection based counting, reflected by the detection score, is highly correlated to the crowd density.
In scenes with sparse crowds, the estimations are reliable, and the detection scores are also higher than those of congested scenes.
On the other hand, in crowded scenes, the corresponding object sizes tend to be very small.
Detection in these scenes is less reliable, leading to low detection scores and recall rates.
Consequently, the predicted crowd counts will be \emph{underestimated}; while the regression based counting methods could perform better on these occasions.
Figure \ref{fig:fig_det_reg}(b) provides the crowd density map visualization on the same image in (a), outputted by a 5-layer CNN based regression network with similar structure employed in  \cite{sam2017switching}.
We find that the estimations in remote  congested areas are quite reasonable. 
However, in background regions near the camera viewpoint, there exist false alarm hot spots on the pavement.
The relationship between ground-truth counts and corresponding predictions is plotted in Figure \ref{fig:fig_det_reg}(d).
Note that the prediction dots for patches with lower ground-truth counts are mostly above the dashed line.
This indicates that prediction counts in these scenes are mostly larger than the ground-truth.
Hence, being not aware of the location of each individual, and directly applying the regression based approaches to low density scenes may lead to \emph{overestimated} results. 

Based on the above ablation analysis, we may find that the detection and regression based counting approaches have their different strengths on different crowd densities.
The regression based method is preferred for congested scenes.
Without localization information for each person, applying them to low density scenes tends to overestimate counts.
The detection based approach could localize and count each person precisely on these occasions since they are expected settings for object detectors.
However, its reliability degenerates in crowded scenes due to small target sizes and occlusion.

\begin{figure}[htbp]
	\centering
	\includegraphics[width=7.0cm]{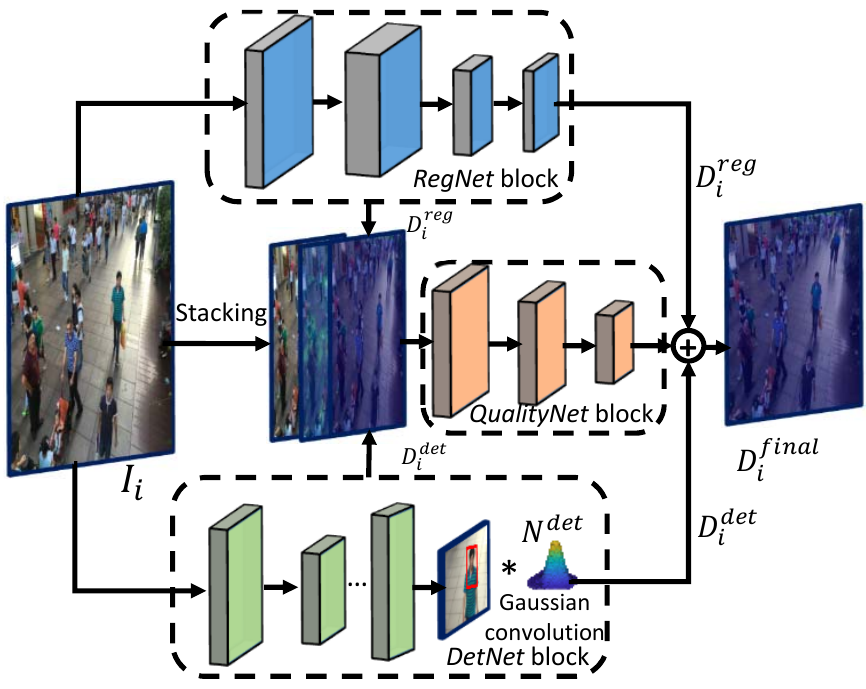}
	\caption{The architecture of our proposed \emph{DecideNet}. Image patches are sent to the \emph{RegNet} and \emph{DetNet} blocks for two types of density maps $D_{i}^{reg}$ and $D_{i}^{det}$ estimation. The final density map $D_{i}^{final}$ is outputted by the \emph{QualityNet}, which adaptively decides the attention weight between two density maps for each pixel. Three blocks are jointly learned on the training data.}
	\label{fig:gen_arch}
\end{figure}

Therefore, we may find that a conventional crowd counting method which only relies on either detection or regression is limited when handling real scenes with unavoidable density variations.
An ideal counting method, on the other hand, should have an \emph{adaptive} ability to choose the appropriate counting mode according to the crowd density:  
in low density scenes, it is expected to count by localizing as an object detector; whereas in congested scenes, it should behave in a regression manner.
Motivated by this understanding, we propose a novel crowd counting framework named as \emph{DecideNet} (\emph{DE}te\emph{C}t\emph{I}on and \emph{D}ensity \emph{E}stimation \emph{Network}), as shown in Figure \ref{fig:gen_arch}.
To the best of our knowledge, \emph{DecideNet} is the first framework, which is capable of perceiving the crowd density for each pixel in a scene and adaptively deciding the relative weights for detection and regression based estimations.

In detail, for a given scene, the \emph{DecideNet} first estimates two kinds of crowd densities maps by detecting individuals and regressing pixel-wise densities, respectively.
To capture the subtle variation in crowd densities, an attention module \emph {QualityNet} is proposed to assess the reliability of two types of density maps with the additional supervision of detection scores.
The final count is obtained under guidance from \emph{QualityNet} to allocate adaptive attention weights for the two density maps.
Parameters in our proposed \emph{DecideNet} are end-to-end learnable by minimizing a joint loss function.

In summary, we make the following contributions:
\begin{itemize}
	\setlength{\itemsep}{0pt}
	\setlength{\parsep}{0pt}
	\setlength{\parskip}{0pt}
	\item We find that real-world crowd counting occasions are frequently faced with great density variations. While existing estimation methods, which either rely exclusively on detection or regression, are unable to provide precise estimations along the whole density range.
	\item Based on the complementary property of two types of crowd counting methods, we design a novel framework \emph{DecideNet}, which can capture this variation and estimate optimal counts by assigning adaptive weights for both detection and regression based estimations.
	\item Experimental results reveal that our method achieves state-of-the-art performance on public datasets with varying crowd densities. 
\end{itemize}

\section{Related works}
\label{sec:related}
\paragraph{Crowd counting by detection.}
Early works addressing the crowd counting problem major follow the counting by detection framework.
Region proposal generators \cite{dollar2012pedestrian,uijlings2013selective} are firstly used to propose potential regions that include persons.
Low-level features \cite{dalal2005histograms,leibe2005pedestrian,sabzmeydani2007detecting,viola2003detecting} are then used for feature representation.
Different binary classifiers including Naive Bayes \cite{chan2009bayesian}, Random Forest \cite{pham2015count} and their variations \cite{xu2016crowd,gall2011hough} are trained with these features. 
The crowd count is the number of positive samples outputted by the classifier on a test image.
Global detection scores are employed to estimate crowd densities and utilized for object tracking in \cite{rodriguez2011density}.
Recent approaches seek the end-to-end crowd counting solution by CNN based object detectors  \cite{girshick2015fast,ren2015faster,redmon2016you,dai2016r} and greatly improve the counting accuracy.
Though detection based crowd counting is successful for scenes with low crowd density, its performance on highly congested environments is still problematic.
On these occasions, usually only partial of the whole objects are visible, posing great challenge to object detectors for localization.
Therefore, part and shape based models are introduced in  \cite{felzenszwalb2010object,lin2001estimation,wu2007detection}, where ensembles of classifiers are built for specific body parts and regions.
Although these methods mitigate the issue in some degree, counting in evident crowded scenes still remains challenging, since objects in those areas are too small to be detected.

\paragraph{Crowd counting by regression.}
Different from counting by detection, counting by regression estimates crowd counts without knowing the location of each person.
Preliminary works employ edge and texture features such as HOG and LBP to learn the mapping from image patterns to corresponding crowd counts \cite{lempitsky2010learning,pham2015count,xu2016crowd}.
Multi-source information is utilized \cite{idrees2013multi} to regress the crowd counts in extreme dense crowd images.
An end-to-end CNN model adopted from AlexNet is constructed \cite{wang2015deep} recently for counting in extreme crowd scenes.
Later, instead of direct regressing the count, the spatial information of crowds are taken into consideration by regressing the CNN feature maps as crowd density maps \cite{zhang2015cross} .
Observing that the densities and appearances of image patches are of large variations, a multi-column CNN architecture is developed for density map regression \cite{zhang2016single}. 
Three CNN columns with different receptive fields are explicitly constructed for counting crowds with robustness to density and appearance changes.
Similar frameworks are also developed in \cite{onoro2016towards}, where a Hydra-CNN architecture is designed to estimate crowd densities in a variety of scenes.
Better performance can be obtained by further exploiting switching structures  \cite{sindagi2017generating,sam2017switching,kumagai2017mixture} or contextual correlations using LSTM \cite{shang2016end}.
Though counting by regression is reliable in crowded settings, without object location information, their predictions for low density crowds tend to be overestimated.
The soundness of such kind of methods relies on the statistical stability of data, while in such scenarios the instance number is too small to help explore the its intrinsic statistical principle.

\section{Crowd Counting by DecideNet}
\label{sec:method}
\subsection{Problem formulation}
\label{sec:prob_form}
Our solution formulates the crowd counting task as a density map estimation problem. 
It requires $N$ training images $I_{1},I_{2},\cdots,I_{N}$ as inputs.
For a specific image $I_{i}$, a collection of $c_{i}$ 2D points $\textbf{P}_{i}^{gt}=\{P_{1},P_{2},\cdots,P_{c_{i}}\}$ is provided by the dataset \cite{chen2012feature,zhang2015cross,zhang2016single}, indicating the ground-truth head positions in the image $I_{i}$.
The ground-truth crowd density map $D^{gt}_{i}$ of $I_{i}$ is generated by convolving annotated points with a Gaussian kernel $\mathcal{N}^{gt}(p|\mu,\sigma^{2}) $ \cite{onoro2016towards}.
Therefore, the density at a specific pixel $p$ of $I_{i}$ could be obtained by considering the effects from all the Gaussian functions centered by annotation points, i.e.,
\begin{equation}
\forall p\in I_{i}, D^{gt}_{i}(p|I_{i})=\sum_{P\in\textbf{P}_{i}^{gt}}\mathcal{N}^{gt}(p|\mu=P,\sigma^{2}).
\label{equ:gt}
\end{equation}
Summing over the density values of all pixels over the entire image $I_{i}$, the total person count $c_{i}$ of $I_{i}$ can be acquired: $\sum_{p\in I_{i}}D^{gt}_{i}(p|I_{i})=c_{i}$.
For a counting model parameterized by $\varOmega$, its objective is to learn a non-linear mapping for $I_{i}$, whereas the difference between the prediction density map $D_{i}^{out}(p|I_{i})$ and the ground-truth $D^{gt}_{i}(p|I_{i})$ is minimized.

Traditional crowd counting by density estimation methods regress density maps by minimizing the pixel-wise Euclidean loss to the ground-truth \cite{wang2015deep,boominathan2016crowdnet,zhang2016single,onoro2016towards}.
However, as we have analyzed in introduction, counting by purely regression would result in the \emph{overestimation} problem on occasions with low crowd densities.
Oppositely, counting by detection works comparably better in those scenes, since low crowd density is the expecting environment to an object detector.

In practical applications, the crowd density varies both spatially and temporally.
Hence, deciding the crowd counts exclusively based on either regression or detection is insufficient.
\emph{DecideNet} is motivated by their complementary property to address this problem.
As shown in Figure \ref{fig:gen_arch}, instead of counting people either by merely regressing density maps, or applying an object detector over the whole image, \emph{DecideNet} simultaneously estimates crowd counting with both detection and regression modules. 
Later, an attention block is utilized to decide which estimation result should be adopted for a specific pixel.
Three CNN blocks are included in our framework: the \emph{RegNet}, the \emph{DetNet} and the \emph{QualityNet}, parameterized by $\varOmega=(\varOmega_{det}, \varOmega_{reg}, \varOmega_{qua})$.
The parameters for three CNN blocks could be jointly learned on the training set.

\subsection{The RegNet block}
\label{sec:reg}
\begin{figure}[htbp]
	\centering
	\includegraphics[width=7.2cm]{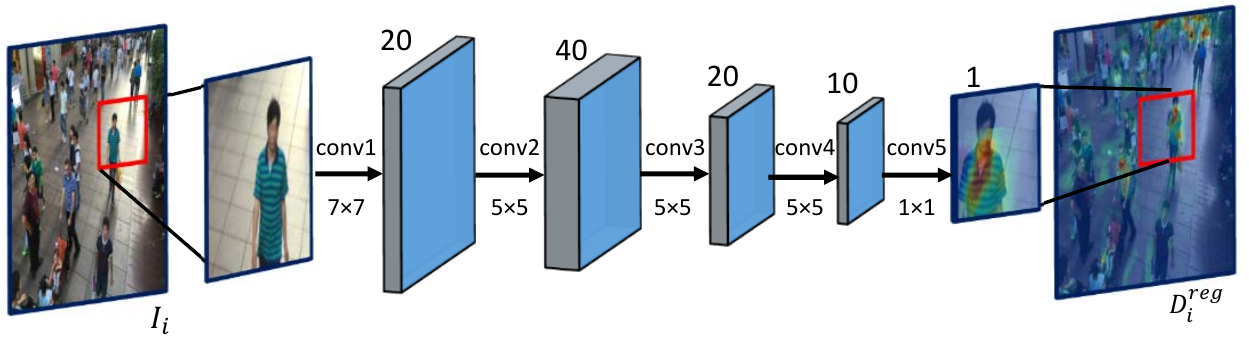}
	\caption{The \emph{RegNet} block consisting of 5 fully convolutional layers. It outputs the crowd density map $D_{i}^{reg}$ of each pixel in image patches without predicting the head locations.}
	\label{fig:regnet}
\end{figure}

The \emph{RegNet} block counts crowds in the absence of localizing each individual.
Without knowing the specific location of each head in the input image patch, it directly estimates the crowd density for all the pixels in $I_{i}$ with a fully convolutional network:
\begin{equation}
\mathcal{F}^{reg}(I_{i}|\varOmega_{reg})=D_{i}^{reg}(p|\varOmega_{reg},I_{i}).
\end{equation}
As shown in Figure \ref{fig:regnet}, the \emph{RegNet} block consists of 5 convolutional layers.
Because it is designed to capture the general crowd density information, larger filters' receptive fields will grasp more contextual details, which is more beneficial for modeling the density maps.
Therefore, in our implemented \emph{RegNet} block, the ``conv1" layer has 20 filters with a $7\times 7$ kernel size.
$40$ filters with a $5 \times 5$ kernel size are set as the ``conv2" layer.
In order to capture scale and orientation invariant person density features, the ``conv1" and ``conv2" layers are followed by two $2 \times 2$ max-pooling layers.
The ``conv3" and ``conv4" layers both have $5 \times 5$ filter sizes with 20 and 10 filters, respectively.
Since the density estimation result could be viewed as a CNN feature map with only one channel, we add a ``conv5" layer with only one filter and a ``$1 \times 1$" filter size. 
This layer is responsible to return the regression based crowd density map $D_{i}^{reg}$, in which value on each pixel represents the estimated count at that point.
A ReLU unit is applied after the ``conv5" layer ensuring that the output density map will not contain negative values.

\subsection{The DetNet block}
\label{sec:det}
\begin{figure}[htbp]
	\centering
	\includegraphics[width=7.2cm]{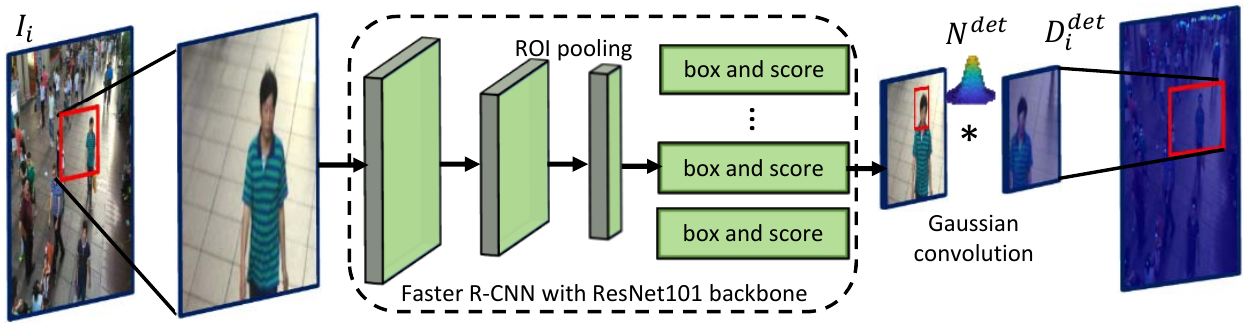}
	\caption{The proposed \emph{DetNet} block is built upon the Faster R-CNN network. A Gaussian convolutional layer is plugged after the bounding box outputs to generate the detection based crowd density map $D_{i}^{det}$.} 
	\label{fig:detnet}
\end{figure}

To handle varying perspectives, crowd densities and appearances, existing density estimation methods  \cite{zhang2015cross,zhang2016single,sam2017switching,kumagai2017mixture} consist of several CNN structures like the \emph{RegNet} block.
However, without the prior knowledge about the exact position of each person in the image patches, the network purely decides the crowd density based on the raw image pixels.
This regression methodology may be accurate in image patches with relatively large crowd densities, while it tends to overestimate the crowd counts in sparse or even ``no-person" (background) scenes.
In our proposed \emph{DecideNet} architecture, the \emph{DetNet} is designed to address this issue by generating the ``location aware" detection based density map $D_{i}^{det}$.
The motivation is intuitive and simple: sparse and non-crowded image patches are expected settings for present CNN based object detectors.
Therefore, compared to use regression networks to count on these patches, using the prior knowledge from outputs of object detectors should substantially relieve the overestimation problem.

The \emph{DetNet} block, illustrated in Figure \ref{fig:detnet} is built based on the above assumption. 
It could be viewed as an extension of the Faster-RCNN network \cite{girshick2015fast} for head detection on the basis of the ResNet-101 architecture.
To be specific, we design a Gaussian convolutional layer and plug it after the bounding box outputs of the original Faster-RCNN network. 
The Gaussian convolutional layer employs a constant Gaussian function $\mathcal{N}^{det}(p|\mu=P,\sigma^{2})$, to convolve over the centers of detected bounding boxes $\textbf{P}_{i}^{det}$ on the original image patch.
The detection based density map $D_{i}^{det}$ is obtained by this layer, i.e.,
\begin{equation}
 D^{det}_{i}(p|\varOmega_{det},I_{i})=\sum_{P\in\textbf{P}_{i}^{det}}\mathcal{N}^{det}(p|\mu=P,\sigma^{2}). 
\label{equ:det_net}
\end{equation}
Since the pixel values of $D_{i}^{det}$ are obtained by considering the impact from the points in detection output $\textbf{P}_{i}^{det}$,  $D_{i}^{det}$ is a ``location aware" density map.
Compared to  $D_{i}^{reg}$ from the output of $RegNet$, responses of $D_{i}^{det}$ are more concentrated on specific head locations.
The difference between them is obvious in $D_{i}^{reg}$ of Figure \ref{fig:regnet} and $D_{i}^{det}$ of Figure \ref{fig:detnet}. 

\subsection{Quality-aware density estimation}
\label{sec:att}
\begin{figure}[htbp]
	\centering
	\includegraphics[width=7.5cm]{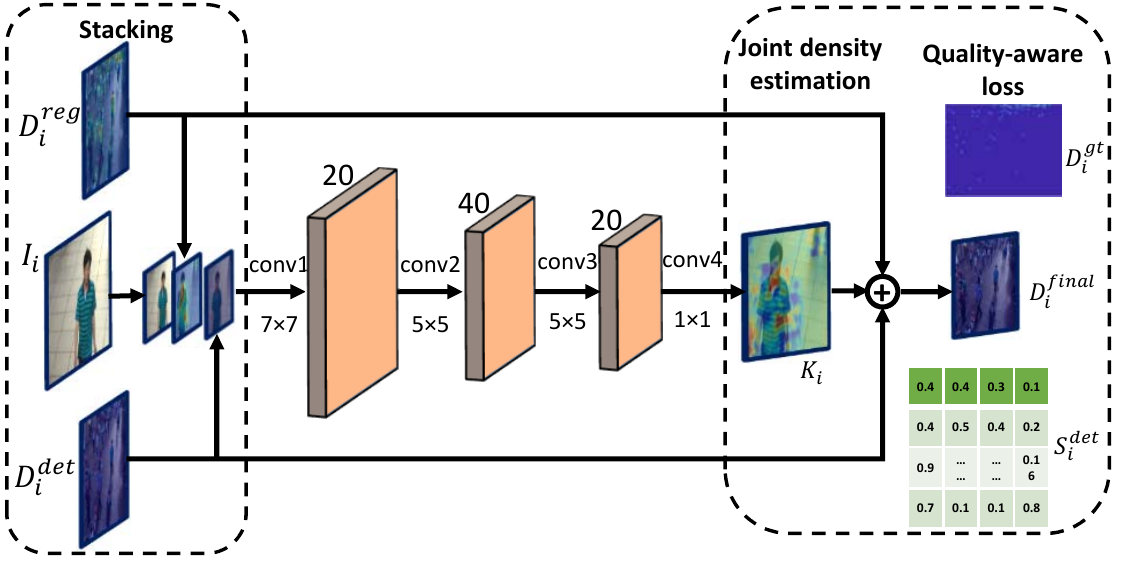}
	\caption{The \emph{QualityNet} block: stacking two density maps and the original image $I_{i}$ as input, it outputs a probabilistic attention map $K_{i}(p|\varOmega_{qua},I_{i})$. The final density estimation $D_{i}^{final}$ is jointly determined by $K_{i}$, $D_{i}^{reg}$ and $D_{i}^{det}$.}
	\label{fig:quanet}
\end{figure}
Herein, we have described the details about obtaining two kinds of density maps: $D_{i}^{reg}$ and $D_{i}^{det}$ for a given image $I_{i}$.
The detection based map $D_{i}^{det}$ employs object detection results for density estimation.
Therefore, it could count persons precisely in sparse density scenes by localizing their head positions.
However, counting via $D_{i}^{det}$ is not accurate on crowded occasions due to the low detection confidence resulted from the small object size and occlusion.
On the contrary, the regression based map $D_{i}^{reg}$, which is unaware of individual locations, is the  preferred estimation for these scenes: the full convolutional network is capable of capturing rich context crowd density information.
Intuitively, one may think that fusing $D_{i}^{reg}$ and $D_{i}^{det}$ by applying average or max pooling \cite{wang2017novel} may obtain better results on varying density crowds.
Nevertheless, even in the same scene, the density may differ significantly in different parts or time intervals.
Therefore, the importance between $D_{i}^{det}$ and $D_{i}^{reg}$ also changes correspondingly for instant pixel values in $I_{i}$.
In \emph{DecideNet}, we propose an attention block \emph{QualityNet}, shown in Figure \ref{fig:quanet} to model the selection process for optimal counting estimations.
It captures the different importance weight of two density maps by dynamically assessing the qualities of them for each pixel.

For a given $I_{i}$, the \emph{QualityNet} block firstly upsamples $D^{det}_{i}$ and  $D_{i}^{reg}$ to the same size of $I_{i}$.
Then $D^{det}_{i}$, $D_{i}^{reg}$ and $I_{i}$ are stacked together as the \emph{QualityNet} input with 5 channels.
Four fully convolutional layers and a pixel-wise sigmoid layer is followed to output a probabilistic attention map $K_{i}(p|\varOmega_{qua},I_{i})$.
We define the specific value of $K_{i}(p|\varOmega_{qua},I_{i})$ at the pixel $p$ reflects the importance of the detection based density map $D_{i}^{det}$, compared to the regression counterpart $D_{i}^{reg}$. 
As a result, the \emph{QualityNet} block could decide the relative \emph{reliability} (i.e., the quality) between $D_{i}^{det}$ and  $D_{i}^{reg}$.
A higher $K_{i}(p|\varOmega_{qua},I_{i})$ at pixel $p$ means a higher attention we should rely on the detection, rather than the regression density estimation for $p$.
Hence, we could further define the final density map estimation $D^{final}_{i}(p|I_{i})$ as a weighted sum between two density maps $D_{i}^{reg}$ and $D_{i}^{det}$, guided by the attention map $K_{i}$:
\begin{equation}
\begin{split}
D^{final}_{i}(p|I_{i})=&K_{i}(p|\varOmega_{qua},I_{i})\odot D^{det}_{i}(p|\varOmega_{det}, I_{i})+ \\
&(\textbf{J}-K_{i}(p|\varOmega_{qua},I_{i})) \odot D_{i}^{reg}(p|\varOmega_{reg},I_{i}),
\end{split}
\end{equation}
whereas $\odot$ is the Hadamard product for two matrices and the $\textbf{J}$ is an all-one-matrix with the same size of $K_{i}$.

\section{Model Learning}
\label{sec:train}
Parameters of \emph{DecideNet} $\varOmega$ consist of three parts: $\varOmega_{reg}$, $\varOmega_{det}$ and $\varOmega_{qua}$.
Hence, we generalize the training process as a multi-task learning problem.
The overall loss function $L_{decide}$, is given by Eq. (\ref{equ:all_loss}):
\begin{equation}
L_{decide}=L_{reg}+L_{det}+L_{qua},
\label{equ:all_loss}
\end{equation}
whereas the $L_{reg}$, $L_{det}$ and $L_{qua}$ are the losses for \emph{RegNet}, \emph{DetNet} and \emph{QualityNet}, respectively.
$L_{decide}$ could be optimized via Stochastic Gradient Descent  with annotated training data.
In each iteration, gradients for $L_{reg}$, $L_{det}$ and $L_{qua}$ are alternatively calculated and employed to update corresponding parameters.
To be specific, for the loss of the \emph{RegNet} component, we employ the pixel-wise mean square error as the loss function. That is:
\begin{equation}
L_{reg}=\dfrac{1}{N}\sum_{i}\sum_{p\in I_{i}}\left[D_{i}^{reg}(p|\varOmega_{reg},I_{i})-D^{gt}_{i}(p|I_{i})\right]^{2},
\end{equation}
whereas $N$ is the total number of training images.

For the $DetNet$ block, different from the regression counterpart, the responses on the density map $D_{i}^{det}$ mostly concentrate on the detected head centers.
Directly minimizing the difference between $D_{i}^{det}$ and $D^{gt}_{i}$ involves in overwhelmed negative pixel samples, i.e., background pixels without head detections.
Hence, instead of using the pixel-wise Euclidean loss as error measurement, we employ the bounding boxes as supervision. 
In this way, optimizing $\varOmega_{det}$ is equivalent to minimizing the classification and localization error in the original Faster R-CNN \cite{ren2015faster}:
\begin{equation}
\begin{split}
L_{det}=\dfrac{1}{N}\sum_{i}  \left[L_{cls}(\textbf{P}_{i}^{det},\textbf{P}_{i}^{gt}|\varOmega_{det})+L_{loc}(\textbf{P}_{i}^{det},\textbf{P}_{i}^{gt}|\varOmega_{det})\right].
\label{equ:det}
\end{split}
\end{equation}
Due to the fact that only the centers of individuals' heads are provided as the annotation on crowd density estimation datasets, we manually label the bounding boxes on partial of the training set points.
Later, we employ the average width and height of them for the bounding box supervision in Eq. (\ref{equ:det}).

The loss function $L_{qua}$ for the attention module \emph{QualityNet} should measure two kinds of errors.
One is the difference between the final crowd density map $D_{i}^{final}$ and the ground-truth density map $D_{i}^{gt}$. 
This error is similar to that we have defined in $L_{reg}$.
The second error measures the quality of the output probabilistic map $K_{i}$ in \emph{QualityNet}.
Recall that $K_{i}(p|\varOmega_{qua},I_{i})$ is the confidence of how reliable the detection result is at pixel $p$ in the image $I_{i}$.
As we analyzed in Figure \ref{fig:fig_det_reg}(c), this confidence could be reflected by the object detection score $S^{det}(p|I_{i})$ at $p$.
Therefore, we employ the Euclidean distances between the probabilistic attention map $K_{i}$ and object detection score map $S^{det}$ as the second error component in $L_{qua}$.
From another perspective, this error could be considered as a regularization term over the \emph{QualityNet} parameters $\varOmega_{qua}$, by incorporating detection scores as prior information.
In experiment evaluation, we will show that this regularization is indispensably beneficial to the performance of our proposed $DecideNet$ architecture.
Since the object detection qualities are brought into this loss function, we name it as the ``quality-aware" loss.
The final formulation of this loss $L_{qua}$ is defined as following:
\begin{align}
L_{qua}=&\dfrac{1}{N}\sum_{i}\sum_{p\in I_{i}}\left\{ \left[D_{i}^{final}(p|\varOmega_{qua},I_{i})-D^{gt}_{i}(p|I_{i})\right]^{2}+ \right. \nonumber \\
& \left. {} \lambda \lVert K_{i}(p|\varOmega_{qua},I_{i})-S^{det}(p|I_{i})\rVert^2 \Big. \right\},
\label{equ:qua_loss}
\end{align}
where $\lambda$ is the hyper-parameter to balance the importance between two errors.

\section{Experimental Results}
\label{sec:exp}
\subsection{Evaluation settings}
\label{sec:eval_set}
Our proposed method is evaluated on three major crowd counting datasets \cite{chen2012feature,zhang2015cross,zhang2016single}  collected from real-world surveillance cameras.
For all datasets, \emph{DecideNet} is optimized with 40k steps of iterations.
We set the initial learning rate at 0.005 and cut it by half in each 10k steps.
Then the best model is selected over the validation data.
Instead of sending the whole image to \emph{DecideNet} during training, we follow the strategy used in \cite{sam2017switching,boominathan2016crowdnet,onoro2016towards} to crop images into $4 \times 3$ patches. 
In this way, the number of samples for training the regression network is boosted.
Each patch is then augmented by random vertical and horizontal flipping with a probability of $0.5$.
We also add uniform noise ranging in $[-5,5]$ on each pixel in the patch with a probability of $0.5$ for data augmentation.
To optimize the parameters for the \emph{RegNet} and the \emph{QualityNet}, the ground-truth density maps are obtained by applying the Gaussian kernel $\mathcal{N}^{gt}(p|\mu,\sigma^{2})$ with $\sigma=4.0$ and a window size of 15.
In each iteration, the object detection score map $S^{det}(p|I_{i})$ is acquired by evaluating $I_{i}$ on the \emph{DetNet}.
For each pixel $p$ in the detected bounding boxes, the value of $S^{det}(p|I_{i})$ is filled with corresponding detection score.
For the rest of pixels which are not included in any bounding boxes, they are filled with a default value set at 0.1. 
The score map is downsampled to the same size of $K_{i}$ in order to calculate the ``quality-aware" loss $L_{qua}$.
We follow the convention of existing works \cite{sindagi2017survey,zhang2015cross,pham2015count} to use the mean absolute error (MAE) and mean squared error (MSE) as the evaluation metric. 
The MAE metric reveals the accuracy of the algorithm for crowd estimation, while the MSE metric indicates the robustness of estimation.

\subsection{The Mall dataset}
\label{sec:mall}
The Mall dataset \cite{chen2012feature} contains 2000 frames, collected in a shopping mall.
Each frame has a fixed resolution of $320 \times 240$.
We follow the pre-defined settings to use the first 800 frames as the training set and the rest 1200 frames as the test set.
The validation set is selected randomly from 100 images in the training set.
We compare our \emph{DecideNet} with both detection based approaches: SquareChn Detector \cite{benenson2014ten}, R-FCN \cite{dai2016r}, Faster R-CNN \cite{ren2015faster}; and regression based approaches: Count Forest \cite{pham2015count}, Exemplary Density \cite{wang2016fast}, Boosting CNN \cite{walach2016learning}, MoCNN \cite{kumagai2017mixture}, Weighted VLAD \cite{sheng2016crowd}.
The evaluation results are exhibited in Table \ref{tab:mall}.
\begin{table}[htbp]
	\scriptsize
	\centering
	\begin{tabular}{|c|c|c|}
		\hline
		\textbf{Method} & \textbf{MAE} & \textbf{MSE} \\
		\hline
		SquareChn Detector \cite{benenson2014ten} &  20.55 & 439.1 \\
		\hline
		R-FCN \cite{dai2016r} & 6.02 & 5.46 \\
		\hline
		Faster R-CNN \cite{ren2015faster} & 5.91 & 6.60 \\
		\hline
		Count Forest \cite{pham2015count} & 4.40   & 2.40 \\
		\hline
		Exemplary Density \cite{wang2016fast} & 1.82  & 2.74 \\
		\hline
		Boosting CNN \cite{walach2016learning} & 2.01  & N/A \\
		\hline
		MoCNN \cite{kumagai2017mixture} & 2.75  & 13.40 \\
		\hline
		Weighted VLAD \cite{sheng2016crowd} & 2.41  & 9.12 \\
		\hline
		\emph{DecideNet} & \textbf{1.52} & \textbf{1.90} \\
		\hline
	\end{tabular}%
	\caption{Comparison results of different methods on the Mall dataset. The MAE and MSE error of our proposed \emph{DecideNet} is significant lower than other approaches. }
	\label{tab:mall}%
\end{table}%

From Table \ref{tab:mall}, we can observe the detection based approaches \cite{benenson2014ten,dai2016r,ren2015faster} generally perform worse than the regression counterparts.
Even the most recent CNN based object detectors \cite{dai2016r,ren2015faster} still have a large performance gap to the CNN based regression approaches \cite{walach2016learning,kumagai2017mixture,sheng2016crowd}.
Our proposed \emph{DecideNet} obtains the minimum error on both MAE and MSE metrics.
Compared to the best approach ``Boosting CNN", which based on regression, \emph{DecideNet} reveals 0.49 point improvement on MAE metric.
This is achieved without using the ensemble scheme employed by the ``MoCNN" and ``Boosting CNN" methods.
Moreover, the MSE metric of the \emph{DecideNet} is merely 1.90.
This is significantly lower than other state-of-the-art methods, which either use detection or regression approach.
This gain rationally results from our density estimations formulated from both detection and regression results.

\subsection{The ShanghaiTech PartB dataset}
\label{sec:shb}
\small
\begin{table}[htbp]
	\scriptsize
	\centering
	\begin{tabular}{|c|c|c|}
		\hline
		\textbf{Method} & \textbf{MAE} & \textbf{MSE} \\
		\hline
		R-FCN \cite{dai2016r} & 52.35 & 70.12 \\
		\hline
		Faster R-CNN \cite{ren2015faster} & 44.51 & 53.22 \\
		\hline
		Cross-scene \cite{zhang2015cross} & 32.00    & 49.80 \\
		\hline
		M-CNN \cite{zhang2016single} & 26.40  & 41.30 \\
		\hline
		FCN \cite{marsden2016fully} & 23.76 & 33.12 \\
		\hline
		Switching-CNN \cite{sam2017switching} & 21.60  & 33.40 \\
		\hline
		CP-CNN \cite{sindagi2017generating} & \textbf{20.1} & 30.1 \\
		\hline
		\emph{DecideNet} & 21.53 & 31.98 \\
		\hline
		\emph{DecideNet+R3} & 20.75 & \textbf{29.42} \\
		\hline
	\end{tabular}%
	\caption{Comparison results of different methods on the ShanghaiTech PartB dataset.}
	\label{tab:shb}%
\end{table}%

We also perform the evaluation experiments on the ShanghaiTech PartB (SHB) \cite{zhang2016single} crowd counting dataset, which is among the largest datasets captured in real outdoor scenes.
It consists of 716 images taken from business streets in Shanghai, in which 400 of them are pre-defined training set and the rest are the test set.
Compared to the Mall dataset, it poses very diverse scene and perspective types over  greatly changing crowd densities. 
We use 50 randomly selected images in the training set for validation.
Since the resolution of each image is $768 \times 1024$, the patches are cropped from the original image with a size of $256 \times 256$ during training.
Our evaluation result and the comparison to other state-of-the-art methods are shown in Table \ref{tab:shb}. 
Due to the large variation in density and object size on the SHB dataset, the detection based approaches \cite{dai2016r,ren2015faster} perform worse than the others relying on regression.
Specifically, the ensemble and fusion strategy is employed by the M-CNN \cite{zhang2016single}, Switching-CNN \cite{sam2017switching}, CP-CNN \cite{sindagi2017generating} in Table \ref{tab:shb}.
Compared to the Mall dataset, the challenging SHB dataset leads to much higher MAE and MSE on all the methods.
Even though, our proposed method (\emph{DecideNet}; \emph{DecideNet+R3}, which trained with an additional R3  stream in Switching-CNN) is very competitive to existing approaches. 

\subsection{The WorldExpo'10 dataset}
\label{sec:world}
The WorldExpo'10 dataset \cite{zhang2015cross} includes 1132 annotated video sequences collected from 103 different scenes in the World Expo 2010 event.
There are a total number of 3980 frames with sizes normalized to $576 \times 720$. 
The patch size we used for training is $144 \times 144 $. 
The training set consists of 3380 frames and the rests are used for testing.
Since the Region Of Interest (ROI) are provided for test scenes (S1-S5), we follow the fashion of previous method \cite{sindagi2017survey} to only count persons within the ROI area.
We use the same metric, namely MAE, suggested by the author \cite{zhang2015cross} for evaluation.
The results of our proposed approach on each test scene and the comparisons to other methods are listed in Table \ref{tab:world}.

\begin{table}[htbp]
	\scriptsize
	\centering
	\begin{tabular}{|c|c|c|c|c|c|c|}
		\hline
		\multirow{2}[4]{*}[6pt]{\textbf{Method}} & \multicolumn{6}{c|}{\textbf{MAE}} \\
		\cline{2-7}
		& \textbf{S1} & \textbf{S2} & \textbf{S3} & \textbf{S4} & \textbf{S5} & \textbf{Ave} \\
		\hline
		Cross-scene \cite{zhang2015cross} & \textbf{2.00}    & 29.50  & 9.70   & \textbf{9.30}   & \textbf{3.10}   & 12.90 \\
		\hline
		M-CNN \cite{zhang2016single} & 3.40   & 20.60  & 12.90  & 13.00    & 8.10   & 11.60 \\
		\hline
		Local\&Global \cite{shang2016end} & 7.80   & 15.40  & 15.30  & 25.60  & 4.10   & 11.70 \\
		\hline
		CNN-pixel \cite{kang2017beyond} & 2.90   & 18.60  & 14.10  & 24.60  & 6.90  & 13.40 \\
		\hline
		Switching-CNN \cite{sam2017switching} & 4.40   & 15.70  & 10.00    & 11.00   & 5.90   & 9.40 \\
		\hline
		\emph{DecideNet} & \textbf{2.00}   & \textbf{13.14}  & \textbf{8.90} & 17.40  & 4.75   & \textbf{9.23} \\
		\hline
	\end{tabular}%
	\caption{Comparison results of different methods on 5 scenes in the WorldExpo'10 dataset.}
	\label{tab:world}%
\end{table}%
From Table \ref{tab:world}, we can notice that our proposed approach achieves an average MAE at 9.23 across all 5 scenes.
This is the best performance among those obtained by all compared methods, revealing 0.17 improvement on the second best ``Switching-CNN" approach.
It is not that significant, because our error on S4 is a little bit higher.
The reason may lie on the fact that people in S4 majorly gather in crowds at remote areas, posing great challenge for the $DetNet$ to output meaningful estimations.
Therefore, the estimation on S4 are mostly relied on the outputs from $RegNet$.
While without the ensemble regression structure, using the $RegNet$ only may not be able to exhibit the superior counting precision.
We can also notice that the prediction counts of different state-of-the-art methods alter considerably on the 5 scenes, revealing different approaches have their own strengths to specific scenes.
However, \emph{DecideNet} obtains three minimum MAE errors when compared to other approaches.
This indicates \emph{DecideNet} having a good generalization ability and prediction robustness on different scenes. 

\begin{table}[htbp]
	\tiny
	\centering
	\begin{tabular}{|c|c|c|c|c|}
		\hline
		\multirow{2}[4]{*}[6pt]{\textbf{Method}} & \multicolumn{2}{c|}{\textbf{MAE}} & \multicolumn{2}{c|}{\textbf{MSE}} \\
		\cline{2-5}
		& \textbf{Mall}  & \textbf{SHB} & \textbf{Mall}  & \textbf{SHB} \\
		\hline
		\emph{RegNet} only & 3.37  & 42.85 & 4.22  & 63.63 \\
		\hline
		\emph{DetNet} only & 4.50   & 44.90  & 5.60   & 73.18 \\
		\hline
		\emph{RegNet}+\emph{DetNet} (Late Fusion) & 3.93  & 38.63 & 4.96  & 65.27 \\
		\hline
		\emph{RegNet}+\emph{DetNet}+\emph{QualityNet} & 1.83  & 24.93 & 2.27  & 41.86 \\
		\hline
		\emph{RegNet}+\emph{DetNet}+\emph{QualityNet} (quality-aware loss) & \textbf{1.52} & \textbf{21.53} & \textbf{1.90} & \textbf{31.98} \\
		\hline
	\end{tabular}%
	\caption{Qualitative results of different \emph{DecideNet} components on the Mall and SHB dataset.}
	\label{tab:ablation}%
\end{table}%

\subsection{Effects of different components in DecideNet}
To analyze effects of each components of the proposed \emph{DecideNet}, we conduct ablation studies on the Mall and SHB dataset.
The qualitative results are listed in Table \ref{tab:ablation}, which shows several interesting observations. 
First, using the estimations exclusively from either the \emph{RegNet} (``\emph{RegNet} only"), or \emph{DetNet} (``\emph{DetNet} only") only obtains fair results on both datasets.
The estimations from the \emph{RegNet} have lower error than the detection counterparts.
This is possibly due to the fact that most of the image regions are with high crowd density on both datasets.
Further, late fusion by averaging two classes of density maps (``\emph{RegNet}+\emph{DetNet} (Late Fusion)") exhibits improvements than ``\emph{RegNet} only" and ``\emph{DetNet} only" on that SHB dataset.
While on the Mall dataset, it only achieves a mediocre result between two kinds of density estimations.
This indicates that direct late fusion is not robust enough to obtain better results across all kinds of datasets.
Second, with \emph{DecideNet}, even training without the object detection scores regularization  (``\emph{RegNet}+\emph{DetNet}+\emph{QualityNet}"), we obtain significant MAE and MSE decrease as compared with those obtained by the previous methods.
Compared to late fusion, it almost decreases the MAE by half on two datasets, revealing the power of the attention mechanism.
Last but not least, adopting the ``quality-aware" loss during training (``\emph{RegNet}+\emph{DetNet}+\emph{QualityNet} (quality-aware loss)"), the MAE and MSE errors are further reduced on two datasets.
In particular, the MSE decreases from 41.86 to 31.98 on SHB dataset: this shows that the loss can substantially increase the prediction stability on challenging datasets with great variations.

\begin{figure}[htbp]
	\centering
	\includegraphics[width=7.2cm]{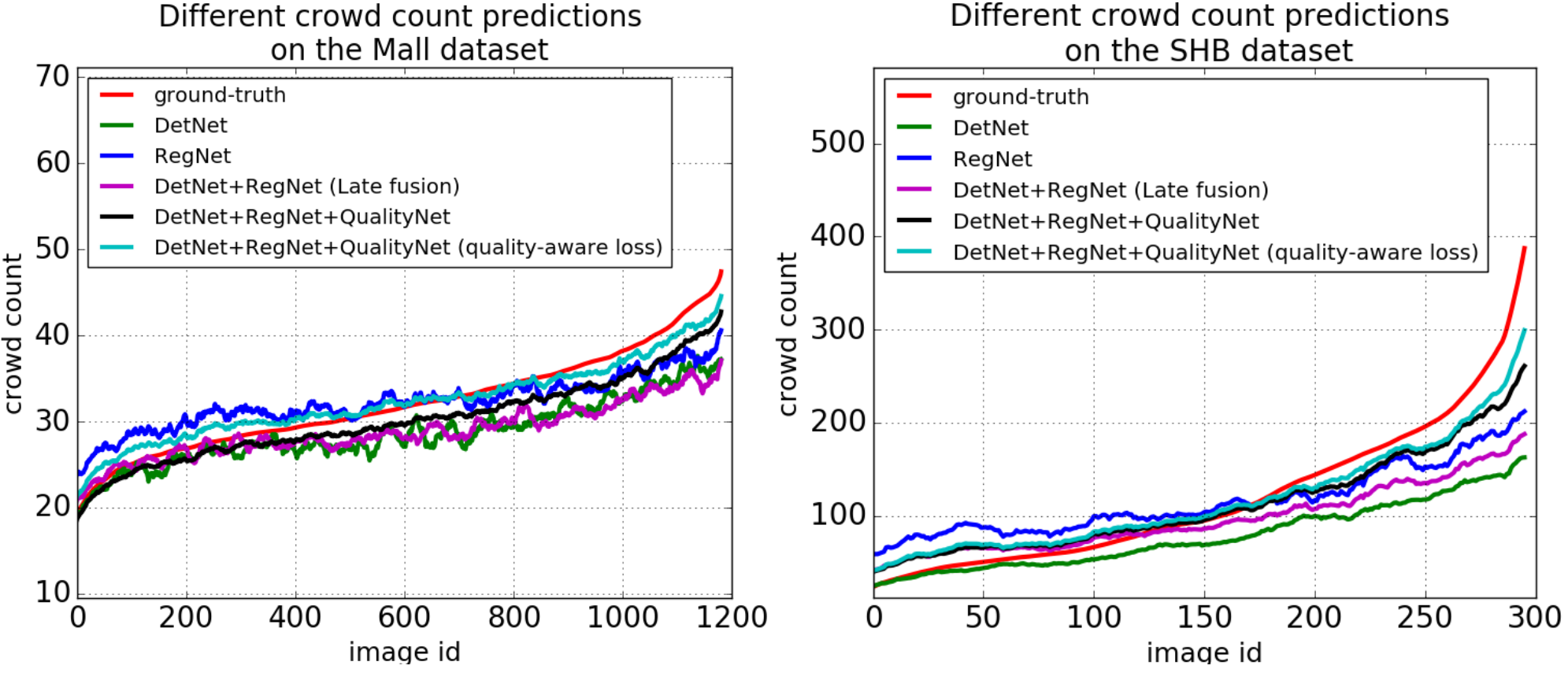}
	\caption{Prediction and the ground-truth crowd counts on the test sets of the Mall (left) and SHB (right) datasets.}
	\label{fig:stat}
\end{figure}

In Figure \ref{fig:stat}, we show the relationships between different crowd count predictions and the ground-truth crowd counts on the test sets of two datasets.
Note that the horizontal axes ``image id" are sorted in ascending order by the number of ground-truth crowd counts.
Clearly, when the numbers of ground-truth crowd count are small, the regression based results from the \emph{RegNet} overestimate the estimations: the blue lines are above the ground-truth red lines in the first half part of the horizontal axis in both figures. 
On the opposite, the detection based result curves (the green lines) fit the red lines well at that region on two datasets. 
However, when the numbers of ground-truth count increase, the estimations of the detection based density map become considerably lower than the red lines, particularly after the second half parts of the horizontal axis. 
The blue lines fit the ground-truth lines best in the middle part of the horizontal axes.
This verifies our observation that regression based estimations are more suitable for high crowded patches. 
Directly applying the late fusion (the purple curves) helps to a certain extent, while its predicted counts are not stable along all images. 
At last, the cyan lines, which represent \emph{DecideNet} outputs, indicate the smallest differences to the ground-truth curves along all parts of the horizontal axes.
That is, the \emph{DecideNet} trained with ``quality-aware" loss exhibits the best estimation results for all kinds of crowd densities on two datasets. 

\subsection{Visualization on density maps}
\begin{figure}[htbp]
	\centering
	\includegraphics[width=7.0cm]{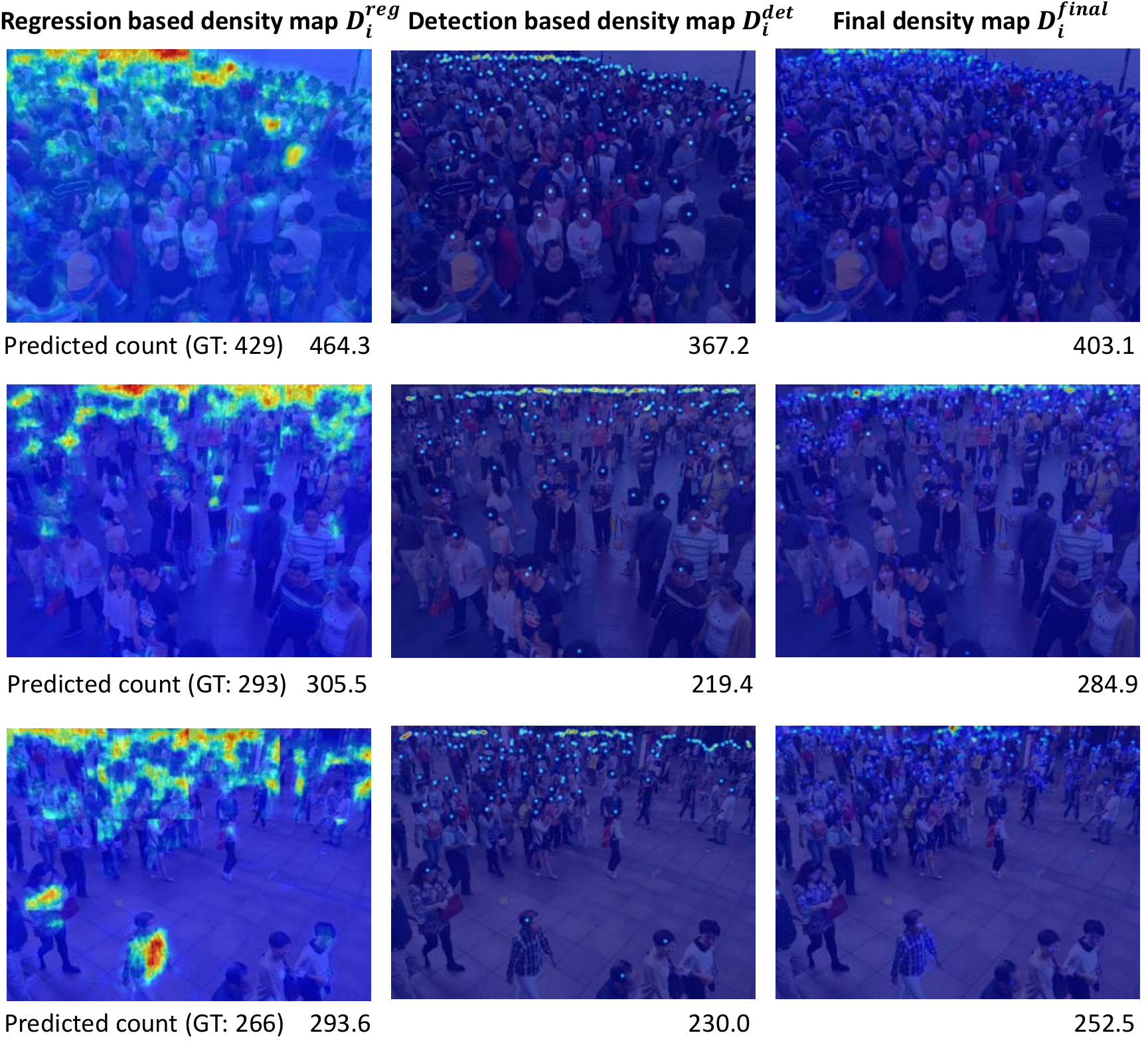}
	\caption{The visualization results of three types of density maps on the SHB dataset (best viewed in color).}
	\label{fig:map}
\end{figure}

To better understand what is learned in our proposed model, we visualize three categories of crowd density maps in the SHB dataset from three blocks: \emph{RegNet}, \emph{DetNet} and \emph{QuialityNet} in Figure \ref{fig:map} (best viewed in color). 

We can discover that the outputs of regression based density maps $D_{i}^{reg}$ (on the most left column) exhibit diffused density estimations along the image regions. 
For the remote areas with highly congested crowds, such predictions from the \emph{RegNet} are reliable.
However, when it comes to the nearby regions with lower crowd density, the results are not satisfactory: some single person bodies are erroneously predicted with very high density.
The prediction counts of the $D_{i}^{reg}$ are also larger than the ground-truth (GT) counts, implying the occurrence of overestimation issue.
Compared to $D_{i}^{reg}$, the detection based density maps $D_{i}^{det}$ (the middle column) are very different: the predicted peak regions are concentrated on the center of heads. 
This is resulted from the fact that these maps are generated from outputs of head detectors. 
We can further observe that the detection based density results are pretty good in nearby low density regions of the given image, while not all the heads are marked with high prediction peaks in the remote areas. 
The underestimated predicted counts of $D_{i}^{det}$ also reflect this phenomenon.
With the attention information from the \emph{QualityNet}, final density maps in the right column reveal very good characteristics: in the nearby region, the estimation prefers the detection results. 
Persons in those areas share very similar estimation patterns with $D_{i}^{det}$.
Oppositely, in remote and congested regions, instead of the ``concentrated dot" patterns, the  density maps are diffused.
\emph{DecideNet} considers the regression based results $D_{i}^{reg}$ are more reliable for those cases.
This confirms that the \emph{QualityNet} block is able to assess the reliability of the corresponding density map value for a specific pixel.

\section{Conclusion}
In this paper, a novel end-to-end crowd counting architecture named \emph{DecideNet} has been proposed.
It is motivated by the complementary performance of detection and regression based counting methods under situations with varying crowd densities.
To the best of our knowledge, \emph{DecideNet} is the first framework to estimate crowd counts via adaptively adopting detection and regression based count estimations under the guidance from the attention mechanism.
We evaluate the framework on three challenging crowd counting benchmarks collected from real-world scenes with high variation in crowd densities.
Experimental results confirm that our method obtains the state-of-the-art performance on three public datasets.

\section{Acknowledgment}
Jiang Liu and Alexander Hauptmann are supported by the Intelligence Advanced Research Projects Activity (IARPA) via Department of Interior/Interior Business Center (DOI/IBC) contract number D17PC00340.  
The U.S. Government is authorized to reproduce and distribute reprints for Governmental purposes notwithstanding any copyright annotation/herein.  
The views and conclusions contained herein are those of the authors and should not be interpreted as necessarily representing the official policies or endorsements, either expressed or implied, of IARPA, DOI/IBC, or the U.S. Government. 
Chenqiang Gao and Deyu Meng are supported by the China NSFC projects with  No.61571071, No.61661166011, No.61721002.

{\small
\bibliographystyle{ieee}
\bibliography{mycite}

\begin{thebibliography}{10}\itemsep=-1pt

\bibitem{benenson2014ten}
R.~Benenson, M.~Omran, J.~H. Hosang, and B.~Schiele.
\newblock Ten years of pedestrian detection, what have we learned?
\newblock {\em CoRR}, abs/1411.4304, 2014.

\bibitem{boominathan2016crowdnet}
L.~Boominathan, S.~S. Kruthiventi, and R.~V. Babu.
\newblock Crowdnet: a deep convolutional network for dense crowd counting.
\newblock In {\em ACM MM}, 2016.

\bibitem{chan2008privacy}
A.~B. Chan, Z.-S.~J. Liang, and N.~Vasconcelos.
\newblock Privacy preserving crowd monitoring: Counting people without people
  models or tracking.
\newblock In {\em CVPR}, 2008.

\bibitem{chan2009bayesian}
A.~B. Chan and N.~Vasconcelos.
\newblock Bayesian poisson regression for crowd counting.
\newblock In {\em ICCV}, 2009.

\bibitem{chen2016videos}
J.~Chen, J.~Liang, H.~Lu, S.-I. Yu, and A.~Hauptmann.
\newblock Videos from the 2013 boston marathon: An event reconstruction dataset
  for synchronization and localization.
\newblock 2016.

\bibitem{chen2012feature}
K.~Chen, C.~C. Loy, S.~Gong, and T.~Xiang.
\newblock Feature mining for localised crowd counting.
\newblock In {\em BMVC}, 2012.

\bibitem{dai2016r}
J.~Dai, Y.~Li, K.~He, and J.~Sun.
\newblock R-fcn: Object detection via region-based fully convolutional
  networks.
\newblock In {\em NIPS}, 2016.

\bibitem{dalal2005histograms}
N.~Dalal and B.~Triggs.
\newblock Histograms of oriented gradients for human detection.
\newblock In {\em CVPR}, 2005.

\bibitem{dollar2012pedestrian}
P.~Dollar, C.~Wojek, B.~Schiele, and P.~Perona.
\newblock Pedestrian detection: An evaluation of the state of the art.
\newblock {\em TPAMI}, 2012.

\bibitem{felzenszwalb2010object}
P.~F. Felzenszwalb, R.~B. Girshick, D.~McAllester, and D.~Ramanan.
\newblock Object detection with discriminatively trained part-based models.
\newblock {\em TPAMI}, 2010.

\bibitem{gall2011hough}
J.~Gall, A.~Yao, N.~Razavi, L.~Van~Gool, and V.~Lempitsky.
\newblock Hough forests for object detection, tracking, and action recognition.
\newblock {\em TPAMI}, 2011.

\bibitem{gao2016people}
C.~Gao, P.~Li, Y.~Zhang, J.~Liu, and L.~Wang.
\newblock People counting based on head detection combining adaboost and cnn in
  crowded surveillance environment.
\newblock {\em Neurocomputing}, 2016.

\bibitem{ge2009marked}
W.~Ge and R.~T. Collins.
\newblock Marked point processes for crowd counting.
\newblock In {\em CVPR}, 2009.

\bibitem{girshick2015fast}
R.~Girshick.
\newblock Fast r-cnn.
\newblock In {\em ICCV}, 2015.

\bibitem{idrees2013multi}
H.~Idrees, I.~Saleemi, C.~Seibert, and M.~Shah.
\newblock Multi-source multi-scale counting in extremely dense crowd images.
\newblock In {\em CVPR}, 2013.

\bibitem{kang2017beyond}
D.~Kang, Z.~Ma, and A.~B. Chan.
\newblock Beyond counting: Comparisons of density maps for crowd analysis
  tasks-counting, detection, and tracking.
\newblock {\em arXiv preprint arXiv:1705.10118}, 2017.

\bibitem{kumagai2017mixture}
S.~Kumagai, K.~Hotta, and T.~Kurita.
\newblock Mixture of counting cnns: Adaptive integration of cnns specialized to
  specific appearance for crowd counting.
\newblock {\em arXiv preprint arXiv:1703.09393}, 2017.

\bibitem{leibe2005pedestrian}
B.~Leibe, E.~Seemann, and B.~Schiele.
\newblock Pedestrian detection in crowded scenes.
\newblock In {\em CVPR}, volume~1, pages 878--885. IEEE, 2005.

\bibitem{lempitsky2010learning}
V.~Lempitsky and A.~Zisserman.
\newblock Learning to count objects in images.
\newblock In {\em NIPS}, 2010.

\bibitem{lin2001estimation}
S.-F. Lin, J.-Y. Chen, and H.-X. Chao.
\newblock Estimation of number of people in crowded scenes using perspective
  transformation.
\newblock {\em IEEE Transactions on Systems, Man, and Cybernetics-Part A:
  Systems and Humans}, 2001.

\bibitem{marsden2016fully}
M.~Marsden, K.~McGuiness, S.~Little, and N.~E. O'Connor.
\newblock Fully convolutional crowd counting on highly congested scenes.
\newblock {\em arXiv preprint arXiv:1612.00220}, 2016.

\bibitem{onoro2016towards}
D.~Onoro-Rubio and R.~J. L{\'o}pez-Sastre.
\newblock Towards perspective-free object counting with deep learning.
\newblock In {\em ECCV}, 2016.

\bibitem{pham2015count}
V.-Q. Pham, T.~Kozakaya, O.~Yamaguchi, and R.~Okada.
\newblock Count forest: Co-voting uncertain number of targets using random
  forest for crowd density estimation.
\newblock In {\em ICCV}, 2015.

\bibitem{redmon2016you}
J.~Redmon, S.~Divvala, R.~Girshick, and A.~Farhadi.
\newblock You only look once: Unified, real-time object detection.
\newblock In {\em CVPR}, 2016.

\bibitem{ren2015faster}
S.~Ren, K.~He, R.~Girshick, and J.~Sun.
\newblock Faster r-cnn: Towards real-time object detection with region proposal
  networks.
\newblock In {\em NIPS}, 2015.

\bibitem{rodriguez2011density}
M.~Rodriguez, I.~Laptev, J.~Sivic, and J.-Y. Audibert.
\newblock Density-aware person detection and tracking in crowds.
\newblock In {\em ICCV}, 2011.

\bibitem{sabzmeydani2007detecting}
P.~Sabzmeydani and G.~Mori.
\newblock Detecting pedestrians by learning shapelet features.
\newblock In {\em CVPR}, 2007.

\bibitem{sam2017switching}
D.~B. Sam, S.~Surya, and R.~V. Babu.
\newblock Switching convolutional neural network for crowd counting.
\newblock In {\em CVPR}, 2017.

\bibitem{shang2016end}
C.~Shang, H.~Ai, and B.~Bai.
\newblock End-to-end crowd counting via joint learning local and global count.
\newblock In {\em ICIP}, 2016.

\bibitem{sheng2016crowd}
B.~Sheng, C.~Shen, G.~Lin, J.~Li, W.~Yang, and C.~Sun.
\newblock Crowd counting via weighted vlad on dense attribute feature maps.
\newblock {\em TCVST}, 2016.

\bibitem{sindagi2017generating}
V.~A. Sindagi and V.~M. Patel.
\newblock Generating high-quality crowd density maps using contextual pyramid
  cnns.
\newblock In {\em ICCV}, 2017.

\bibitem{sindagi2017survey}
V.~A. Sindagi and V.~M. Patel.
\newblock A survey of recent advances in cnn-based single image crowd counting
  and density estimation.
\newblock {\em Pattern Recognition Letters}, 2017.

\bibitem{uijlings2013selective}
J.~R. Uijlings, K.~E. Van De~Sande, T.~Gevers, and A.~W. Smeulders.
\newblock Selective search for object recognition.
\newblock {\em IJCV}, 2013.

\bibitem{viola2003detecting}
P.~Viola, M.~J. Jones, and D.~Snow.
\newblock Detecting pedestrians using patterns of motion and appearance.
\newblock In {\em ICCV}, 2003.

\bibitem{walach2016learning}
E.~Walach and L.~Wolf.
\newblock Learning to count with cnn boosting.
\newblock In {\em ECCV}, 2016.

\bibitem{wang2015deep}
C.~Wang, H.~Zhang, L.~Yang, S.~Liu, and X.~Cao.
\newblock Deep people counting in extremely dense crowds.
\newblock In {\em ACM MM}, 2015.

\bibitem{wang2017novel}
L.~Wang, C.~Gao, J.~Liu, and D.~Meng.
\newblock A novel learning-based frame pooling method for event detection.
\newblock {\em Signal Processing}, 2017.

\bibitem{wang2016fast}
Y.~Wang and Y.~Zou.
\newblock Fast visual object counting via example-based density estimation.
\newblock In {\em ICIP}, 2016.

\bibitem{wu2007detection}
B.~Wu and R.~Nevatia.
\newblock Detection and tracking of multiple, partially occluded humans by
  bayesian combination of edgelet based part detectors.
\newblock {\em IIJCV}, 2007.

\bibitem{xu2016crowd}
B.~Xu and G.~Qiu.
\newblock Crowd density estimation based on rich features and random projection
  forest.
\newblock In {\em WACV}, 2016.

\bibitem{zhang2015cross}
C.~Zhang, H.~Li, X.~Wang, and X.~Yang.
\newblock Cross-scene crowd counting via deep convolutional neural networks.
\newblock In {\em ICCV}, 2015.

\bibitem{zhang2016single}
Y.~Zhang, D.~Zhou, S.~Chen, S.~Gao, and Y.~Ma.
\newblock Single-image crowd counting via multi-column convolutional neural
  network.
\newblock In {\em CVPR}, 2016.

\bibitem{zhao2008segmentation}
T.~Zhao, R.~Nevatia, and B.~Wu.
\newblock Segmentation and tracking of multiple humans in crowded environments.
\newblock {\em ITPAMI}, 2008.

\end{thebibliography}
}

\end{document}